\documentclass{article}
\usepackage[utf8]{inputenc}
\usepackage{spconf}
\usepackage{float}
\usepackage{graphicx}
\usepackage{multirow}
\usepackage{xcolor}
\usepackage{hyperref}
\usepackage{acronym}
\usepackage{supertabular}
\usepackage{subcaption}
\DeclareRobustCommand{\IEEEauthorrefmark}[1]{\smash{\textsuperscript{\footnotesize #1}}}
\usepackage{enumitem}

\title{PQLM - Multilingual Decentralized Portable Quantum Language Model}

\name{
\begin{tabular}{@{}c@{}}
Shuyue Stella Li$^{\star}$\IEEEauthorrefmark{1} \qquad 
Xiangyu Zhang$^{\star}$\IEEEauthorrefmark{1} \qquad 
Shu Zhou\IEEEauthorrefmark{3}\qquad
Hongchao Shu\IEEEauthorrefmark{1}\\
Ruixing Liang\IEEEauthorrefmark{1} \qquad 
Hexin Liu\IEEEauthorrefmark{4}\qquad
Leibny Paola Garcia\IEEEauthorrefmark{1,}\IEEEauthorrefmark{2}
\end{tabular}}
\address{
\IEEEauthorrefmark{1}Center for Language and Speech Processing, Johns Hopkins University \\
\IEEEauthorrefmark{2}Human Language Technology Center of Excellence, Johns Hopkins University \\
\IEEEauthorrefmark{3}Department of Physics, Hong Kong University of Science and Technology\\
\IEEEauthorrefmark{4}School of Electrical and Electronic Engineering, Nanyang Technological University
}

\begin{document}
%
\maketitle
%
\begingroup
\def\thefootnote{$\star$}
\footnotetext{Equal contribution in alphabetical order}
\endgroup
\newacro{DP}[DP]{Differential Privacy}
\newacro{lstm}[LSTM]{Long-Short Term Memory}
\newacro{vqc}[VQC]{Variational Quantum Classifiers}
\newacro{RNN}[RNN]{Recurrent Neural Network}
\newacro{NISQ}[NISQ]{Noisy intermediate-scale quantum}
\newacro{PLM}[PLM]{pre-trained language models}
\newacro{qlstm}[Q-LSTM]{Quantum \ac{lstm}}
\newacro{qlm}[PQLM]{quantum language model}
\newacro{sa}[SA]{setiment analysis}

\begin{abstract}
With careful manipulation, malicious agents can reverse engineer private information encoded in pre-trained language models. Security concerns motivate the development of quantum pre-training.
In this work, we propose a highly portable \ac{qlm} that can easily transmit information to downstream tasks on classical machines.
The framework consists of a cloud \ac{qlm} built with random Variational Quantum Classifiers (VQC) and local models for downstream applications. We demonstrate the ad hoc portability of the quantum model by extracting \textit{only} the word embeddings and effectively applying them to downstream tasks on classical machines.
Our \ac{qlm} exhibits comparable performance to its classical counterpart on both intrinsic evaluation (loss, perplexity) and extrinsic evaluation (multilingual sentiment analysis accuracy) metrics. We also perform ablation studies on the factors affecting \ac{qlm} performance to analyze model stability.
Our work establishes a theoretical foundation for a portable quantum pre-trained language model that could be trained on private data and made available for public use with privacy protection guarantees.
\end{abstract}


\vspace{-3mm}
\section{Introduction}\vspace{-3mm}
A competitive language model can be extremely useful for downstream tasks such as machine translation and speech recognition despite the domain mismatch between pre-training and downstream tasks \cite{bommasani2021opportunities, wu2021yuan}. They become more powerful with increased training data, but there is a trade-off between data privacy and utility\cite{shi2022selective}. Previous works on ethical AI have shown that \ac{PLM}s memorizes training data in addition to learning about the language \cite{carlini2019secret, carlini2021extracting}, which opens up vulnerabilities for potential adversaries to recover sensitive training data from the model.

\begin{figure}
\centering
\includegraphics[width=0.42\textwidth]{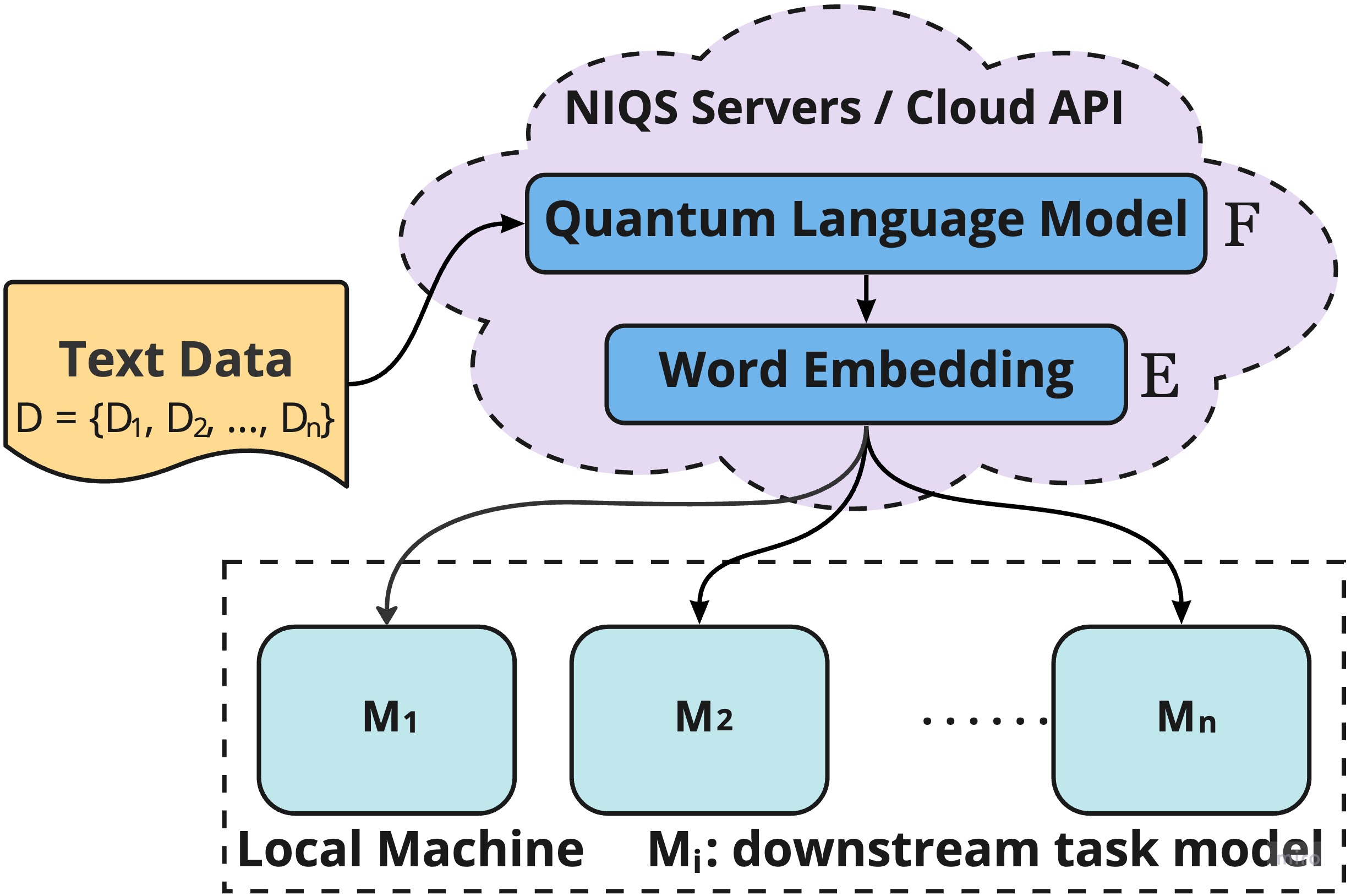}\vspace{-2mm}
\caption{Decentralized Quantum Language Model Pipeline.\\Text data is trained on language model on NISQ servers, the word embeddings are transferred to downstream models $\mathcal{M}_i$}\label{fig:model_pipeline}\vspace{-5mm}
\end{figure}

While some argue that language models should only be trained on data \textit{explicitly} produced for public use \cite{brown2022does}, private data are richer in certain domains compared to public corpora, including dialogue systems, code-mixing languages, and medical applications \cite{zaib2020short, reitmaier2022opportunities}. Therefore, it is essential to develop new methods to mitigate potential data security and privacy problems while being able to take advantage of the rich linguistic information encoded in private data.

Recently, there has been growing interest in leveraging random quantum circuits in neural models to solve data privacy issues \cite{watkins2023quantum}. The entanglement of states from the random configuration of gates in the quantum circuits makes it possible to securely encode sensitive information contained in training data \cite{shaffer2014irreversibility, watkins2023quantum}. The combination of random quantum circuits and decentralized training ensures privacy \cite{yang2021decentralizing}. Additionally, quantum computing has become the next logical step in the development of deep learning for its efficiency in manipulating large tensors \cite{lloyd2013quantum}. 

The architecture of a large quantum computer is vastly different from the classical computer as physical and environmental constraints cannot be met \cite{martinis2021quantum}, which means that the model trained on the large quantum computer is difficult to be directly used by others on the classical computer. Ad hoc portability is defined as the model's ability to transmit the most essential information contained in the language model across different machines.
In this work, we provide a method to seamlessly transmit the information learned from the quantum training step to the classical machine without requiring any additional model adaptations, making the quantum model highly portable.

As shown in Figure \ref{fig:model_pipeline}, we propose a decentralized \ac{qlm} pipeline where private data is fed into a quantum model composed of a \ac{RNN} with its gates replaced with \ac{vqc}\cite{griol2021variational}. As we will describe more in detail in Section \ref{sec:relatedwork}, \ac{NISQ} computers on quantum servers are used for such variational quantum algorithm computations \cite{preskill2018quantum}. 
After the \ac{qlm} hosted on remote \ac{NISQ} is trained to convergence, instead of downloading the entire model to a classical system, we directly use the embeddings trained by the \ac{qlm} to initialize downstream tasks.\\
In summary, our contributions include the following:
\begin{itemize}[noitemsep,topsep=1pt,leftmargin=10pt]
    \item We propose a decentralized pipeline for transmitting knowledge learned from a secure, fast quantum pre-trained model to classical machines for downstream tasks.
    \item We demonstrate the ad hoc portability of our pre-trained language model by showing that extracting the embeddings trained by the \ac{qlm} is sufficient for downstream tasks such as \ac{sa}.
    \item We show the stability of the model across different orders of complexity with ablation studies on factors including the number of qubits and training corpus size.
\end{itemize}

\section{Related Work}\vspace{-2mm}
\textbf{Differential Privacy and Federated Learning\hspace{3mm}}
Traditional approaches to protecting sensitive data such as \ac{DP} introduces random noise to the system to protect individual data change, but usually sacrifices model performance and efficiency \cite{shi2022selective} and makes strong assumptions on the training data \cite{brown2022does}.
More recent work towards a safe data pipeline involves decentralized training for federated training, where the training data is strictly kept to remote machines, and the gradients are exported and aggregated on another machine for downstream tasks \cite{mammen2021federated, li2020federated, gong2020survey}. This resolves the need for any private data to join a centralized data pool from the root. Even then, it is possible to leak sensitive information \cite{bonawitz2017practical}. 

\vspace{3mm}
\noindent\textbf{Quantum ML with Variational Quantum Circuits\label{sec:relatedwork}\hspace{3mm}}
Some recent works attempt to make use of the irrecoverability of quantum circuits in differential privacy algorithms \cite{watkins2023quantum} or federated learning architectures \cite{chen2021federated} involving the use of \ac{vqc}s on \ac{NISQ} clusters with reliable optimization \cite{cerezo2021variational}. 
\ac{vqc}s are quantum circuits with quantum parameters that can absorb the noise inherently contained in quantum computations and can be optimized iteratively with classical gradient descent \cite{chen2022quantum, zhou2020quantum}.
As shown in Figure \ref{fig:vqc}, there are three parts in the \ac{vqc} architecture: (1) the encoding stage where the input vector is encoded; (2) the quantum circuit stage where entanglement strategies are applied with quantum gates and parameters are stored and trained; and (3) the measurement stage where a hermitian operator projects the quantum states onto its eigenvectors \cite{griol2021variational}. 

For decentralized training in which the data and the quantum portion are hosted on a \ac{NISQ} server, any adversaries will not be able to recover the structure or data without knowing the quantum gate configurations in the random circuit, therefore providing a security guarantee \cite{li2019quantum, yang2021decentralizing, shaffer2014irreversibility, preskill2018quantum}. More secure procedures include the Quantum Circuit Obfuscation method that add dummy CNOT gates to the circuit so the data is protected from both the quantum and local machines, yet at the cost of additional computation \cite{suresh2021quantum}.

\vspace{3mm}
\noindent\textbf{Decentralized Quantum Learning Applications\hspace{3mm}}
Some previous work has looked into the area of sequential data modeling \cite{chen2022quantum} and natural language processing \cite{basile2017towards}, but with a focus on the speedup of quantum computations. Applications of quantum decentralized privacy protection approach have been used for speech feature extraction \cite{yang2021decentralizing}, image recognition \cite{qi2022federated}, language processing \cite{yang2022bert}, and reinforcement learning \cite{chen2020variational}. In this paper, we take a single-party delegated training approach that can be easily extended to multi-machine decentralized learning to train a secure \ac{qlm}. We focus on providing a portable transfer of information from the quantum server to the classical side.

\section{Methodology}\vspace{-1mm}
\subsection{Quantum-LSTM Language Model}
\begin{figure}
     \centering
     \begin{subfigure}[b]{0.228\textwidth}
         \centering
         \includegraphics[width=\textwidth]{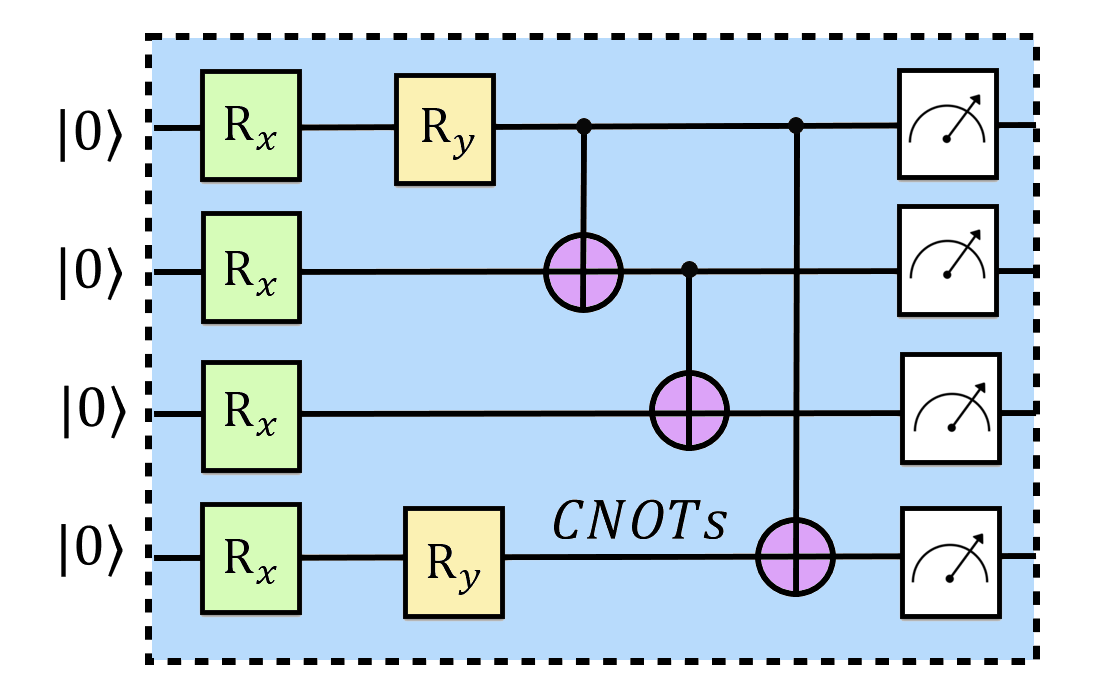}\vspace{-1mm}
         \caption{Quantum Circuit \cite{yang2022bert}}
         \label{fig:vqc}
     \end{subfigure}
     \begin{subfigure}[b]{0.247\textwidth}
         \centering
         \includegraphics[width=\textwidth]{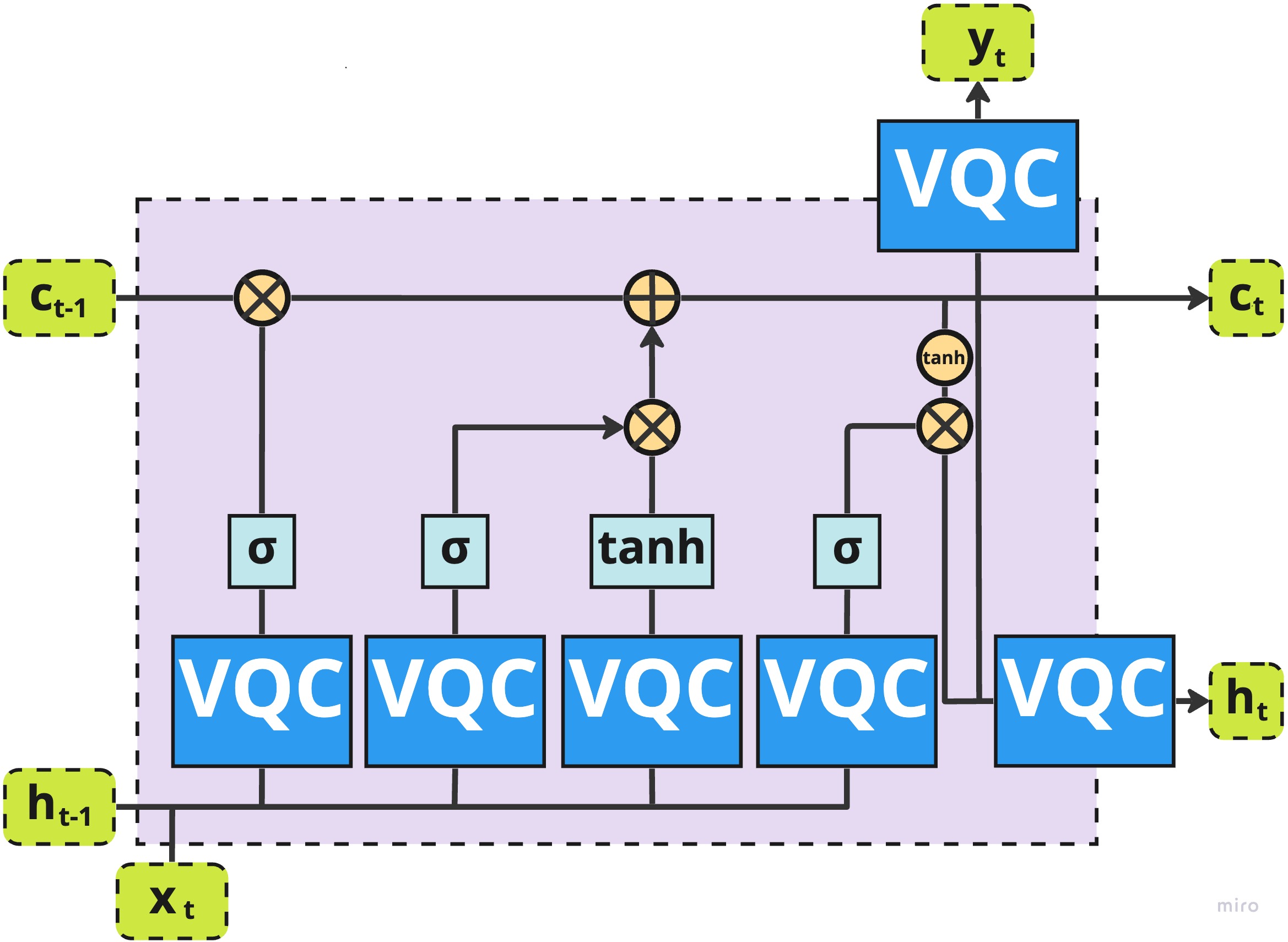}\vspace{-1mm}
         \caption{\ac{qlstm}}
         \label{fig:lstm}
     \end{subfigure}\vspace{-1mm}
     \caption{Model Architecture}\vspace{-4mm}
     \label{fig:model_arch}\vspace{-2mm}
\end{figure}

\vspace{-2mm}
The primary task of language modeling is that given sequence words $x^{1},x^{2},...,x^{t}$, we need to predict the probability $P(x^{t+1} \mid x^{1},x^{2},...x^{t})$. In previous studies, \ac{lstm} has shown to be a good deep learning model for building language models~\cite{zaremba2014recurrent}. In our studies, We use \ac{qlstm} ~\cite{chen2022quantum} which is based on \ac{lstm} model and random \ac{vqc} to train our language model. The basic architecture of \ac{qlstm} is shown in Figure \ref{fig:lstm}. \ac{qlstm} replaces some components such as forget gate, input gate, update gate, and output gate in classical \ac{lstm} with \ac{vqc} and uses the mechanism of backpropagation to update parameters of the \ac{qlstm} model. To meet the coherence time specification, our \ac{qlstm} language model is built with a shallow circuit of 2 layers and each VQC gate is built with 4 qubits. Both the classical and \ac{qlstm} have a word embedding size of 64 and vocab size as the output size. 

As shown in Figure \ref{fig:vqc}, we use a random circuit to encode the rotational vectors to protect the parameters against any 3rd party attacks in the entanglement stage.

\vspace{-2mm}
\subsection{Decentralized Training}\vspace{-2mm}
Quantum computers are in theory exponentially faster than classical computers, making them ideal for training large-scale models. Therefore, decentralized training on a quantum machine will provide both security and speedup. However, because the quantum model learns in the Hilbert space represented by qubits, it will be difficult to load the entire model into classical machines for downstream tasks the same way we do for current pre-trained language models like BERT. We introduce a decentralized \ac{qlm} framework as shown in Figure \ref{fig:model_pipeline}, which simplifies the transmission of information from the quantum side to the classical side using extracted word embeddings.
Given decentralized \ac{qlm} $F$, we input a set of text documents $D_1, D_2, D_3...,D_n$ into the \ac{NISQ} servers. After training, the word embedding $E$ will be extracted from \ac{qlm} and using in local downstream task.\vspace{-2mm}
\begin{equation}
    E = F(D_1, D_2, D_3...,D_n)
    \label{embedding}\vspace{-2mm}
\end{equation}
In Equation \ref{embedding}, the \ac{qlm} $F$ can be any quantum deep learning model that can be used to train the language model.
In this paper, we present \ac{qlstm} as an example to train our language model.  The output word embeddings $E$ is a set of vectors that can be further processed by other classical or quantum routines. Finally, we make our work open source for future explorations \footnote{\url{git@github.com:stellali7/quantunLM.git}}.

\vspace{-3mm}\section{EXPERIMENTS}
\vspace{-4mm}
\subsection{Dataset}\label{sec:dateset}\vspace{-2mm}
We use two distinct datasets to train two language models. The first one is a multilingual Twitter dataset \footnote{\url{https://www.kaggle.com/datasets/jp797498e/twitter-entity-sentiment-analysis}} consisting of 69491 training documents and 998 test documents with 4 different labels (negative, positive, neutral and irrelevant). The second one is an English-Hindi code-mixed Twitter dataset from SemEval-2020 Task 9 \cite{patwa2020semeval}\footnote{\url{https://ritual-uh.github.io/sentimix2020/}}, which consists of 15000 training and validation documents and 3,789 test documents. These two datasets are selected because they are linguistically complex in nature, and they have gold labels for \ac{sa}, which we later use in our evaluation step to train a classification model without introducing further supervision or noise.

\vspace{-2mm}
\subsection{Preprocessing}\vspace{-2mm}
A coarse filtering of the training datasets is performed before they are fed into the language model. First, empty strings, hash symbols, and URLs are removed from the text. All emojis and emoticons are replaced by their English descriptions using the emoji library\footnote{\url{https://pypi.org/project/emoji/}}. Sentiment labels are ignored during language model training.

\vspace{-2mm}
\subsection{Q-LSTM LM vs. Classical LSTM LM}\vspace{-2mm}
In order to fairly assess the performance of the \ac{qlm}, we create a classical language model with the same model architecture and size. With a hidden size of 5 in the classical \ac{lstm}, the number of parameters is of the same magnitude as the \ac{qlstm} with 4 qubits \cite{chen2022quantum}.

\begin{figure}[h!]\vspace{-2mm}
     \centering
     \begin{subfigure}[b]{0.23\textwidth}
         \centering
         \includegraphics[width=\textwidth]{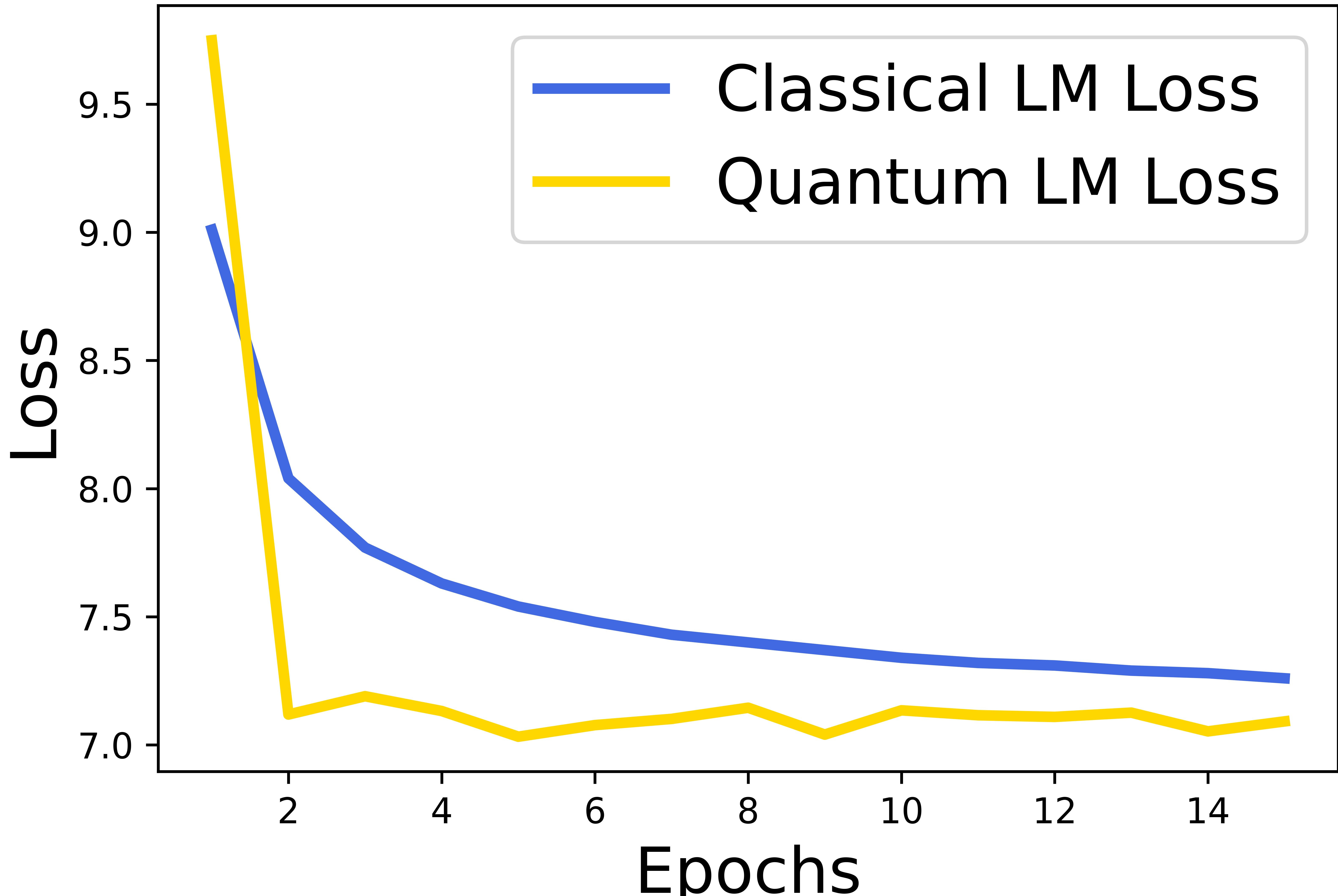}
         \caption{Classical vs. Quantum LM}
         \label{fig:main_loss}
     \end{subfigure}
     \hfill
     \begin{subfigure}[b]{0.23\textwidth}
         \centering
         \includegraphics[width=\textwidth]{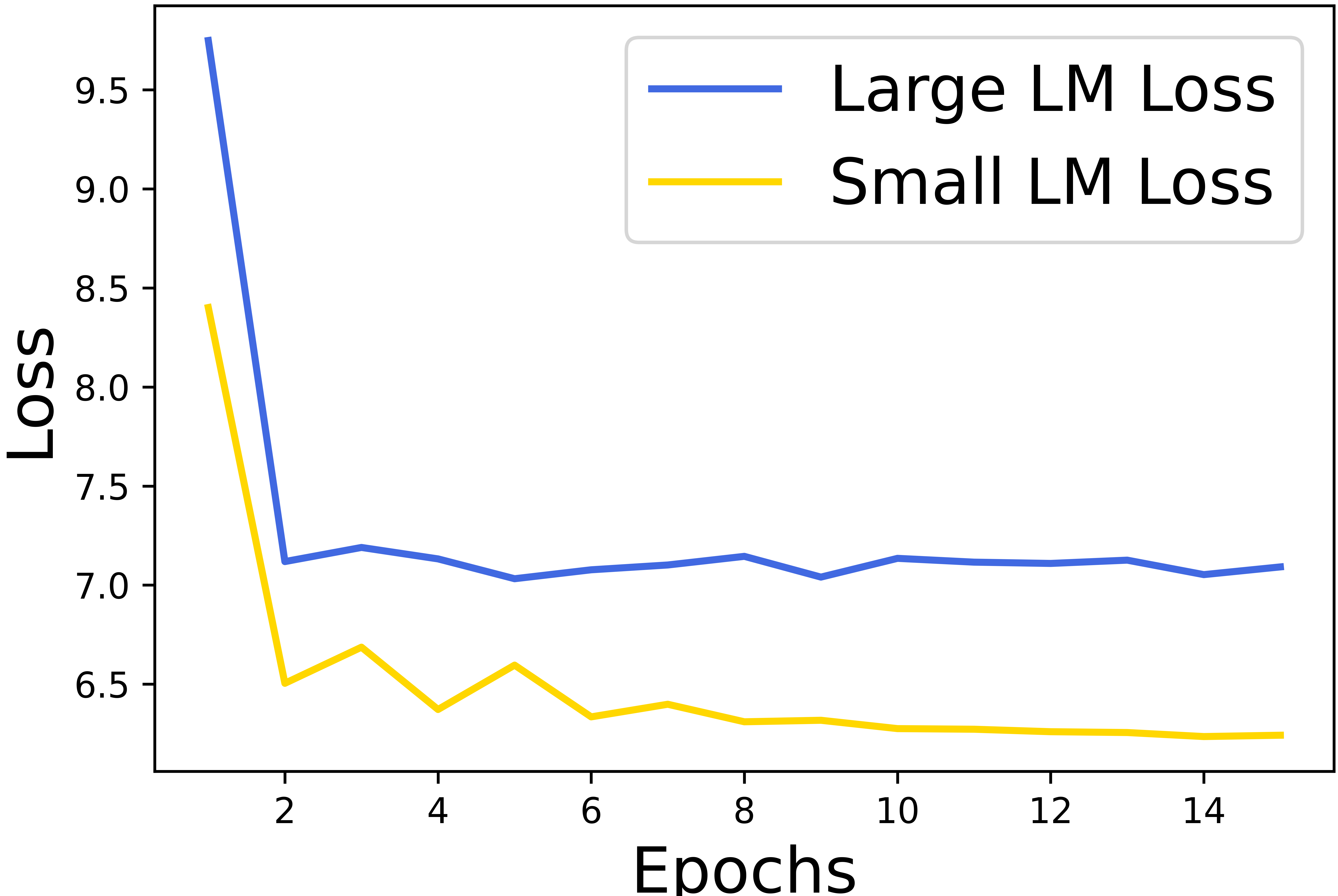}
         \caption{Effect of Dataset Size}
         \label{fig:datasize}
     \end{subfigure}
     \hfill
     \begin{subfigure}[b]{0.23\textwidth}
         \centering
         \includegraphics[width=\textwidth]{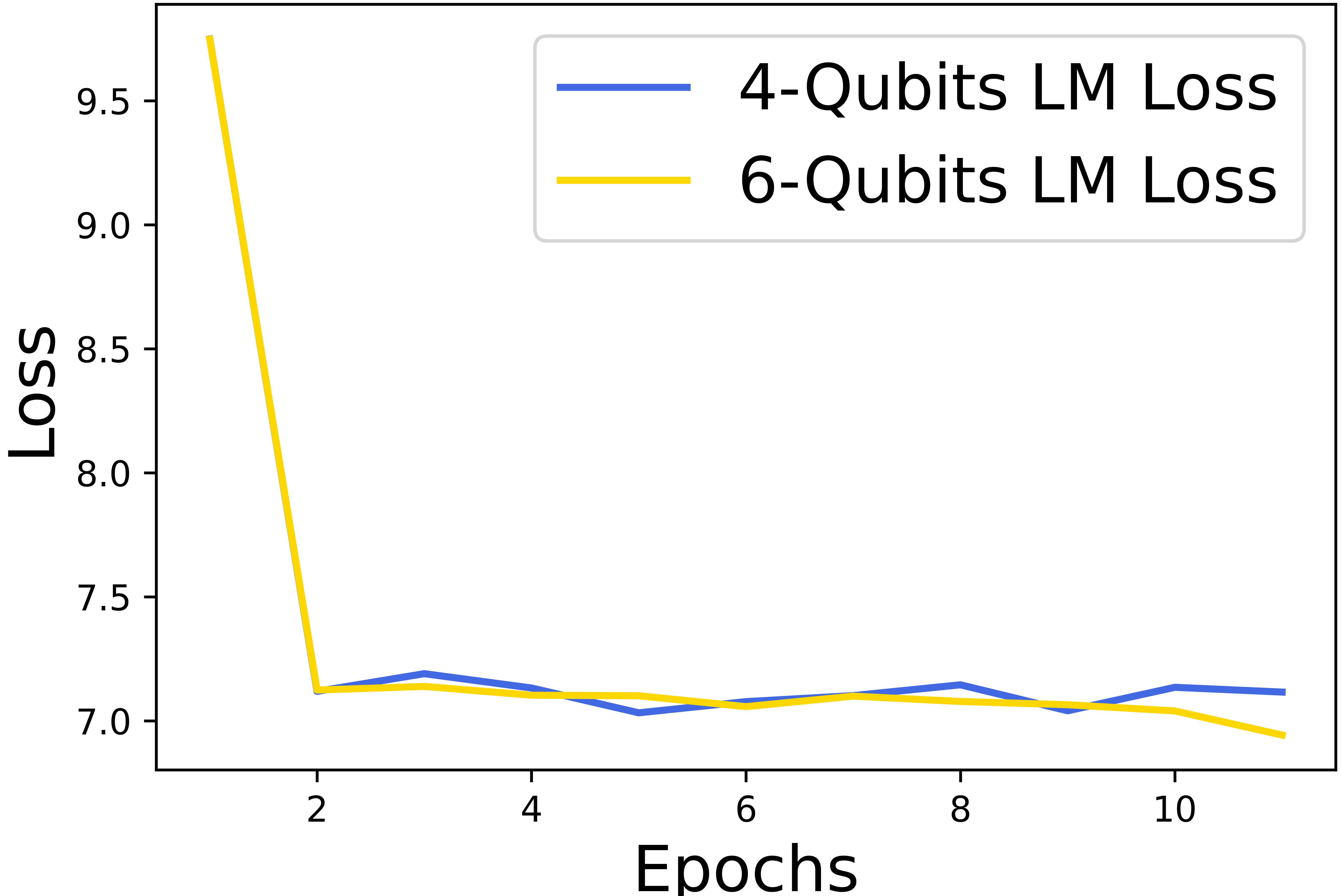}
         \caption{Effect of \# qubits (12 epoch)}
         \label{fig:4v6_all}
     \end{subfigure}
     \hfill
     \begin{subfigure}[b]{0.23\textwidth}
         \centering
         \includegraphics[width=\textwidth]{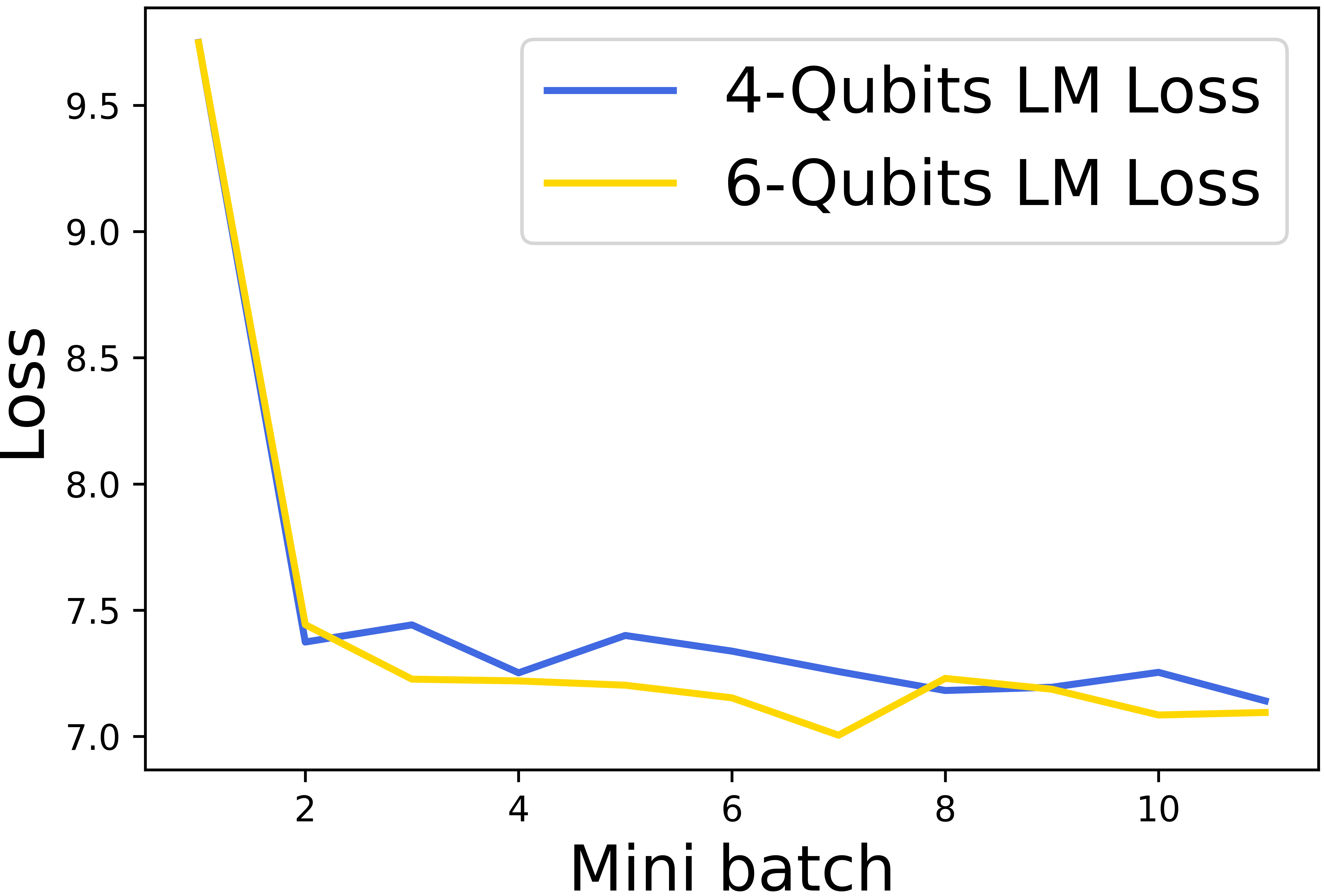}
         \caption{Effect of \# qubits (1 epoch)}
         \label{fig:4v6_1st}
     \end{subfigure}
     \caption{Training Loss for Different Experiments}\vspace{-4mm}
     \label{fig:loss_all}
\end{figure}

The models are trained until the loss converges. The negative log-likelihood loss during training and the model perplexity are recorded to evaluate the model. Figure \ref{fig:main_loss} shows the training loss over 15 epochs of the \ac{qlstm} and its classical counterpart. The \ac{qlm} converges much faster than the classical language model of the same size.\vspace{-2mm}

\begin{table}[ht!]
\centering
\begin{tabular}{c|ccc} 
     \hline
     Model & \ac{lstm} & \ac{qlstm}-4q & \ac{qlstm}-6q \\
     \hline
     perplexity & 1152.78 & 1153.67 & 972.44 \\ 
     \hline
\end{tabular}\vspace{-2mm}
\caption{\label{tab:perp} Model Perplexity on Multilingual Twitter Corpus\\\ac{lstm}: classical \ac{lstm}; \ac{qlstm}-4q: \ac{qlstm} built with 4-qubit \ac{vqc}; \ac{qlstm}-6q: \ac{qlstm} built with 6-qubit \ac{vqc}.}\vspace{-3mm}
\end{table}

Despite the quantum model converging faster than the classical language model, the model quality as measured by model perplexity for both models is extremely close as shown in Table \ref{tab:perp}. Given the computing constraints, we only used 4 qubits to train the \ac{lstm} model, resulting in an extremely limited number of parameters (around 200) for both models. Although the model perplexity is not ideal compared to conventional \ac{PLM}s, it still provides a valuable basis for comparing the classical language model and the \ac{qlm}. Another factor that contributes to the high perplexity of our language model is the multilinguality of the training data, which contains richer linguistic information, making it difficult for the small model to learn. 

\vspace{-2mm}
\subsection{Model Evaluation - Sentiment Analysis}\vspace{-2mm}
We use a downstream \ac{sa} task to compare the quality of the \ac{qlm} with its classical counterpart. Given a sequence of tweets $D_1, D_2...D_n$, we need to predict the sentiment class (positive, neutral, negative, or irrelevant). We train a local transformer-based four-way classifier using the pre-trained word embeddings from the \ac{qlstm} language model. We use 4 transformer blocks and each transformer model has 4 attention heads. In order to avoid introducing more noise to our model pipeline, we use a random subset of the training data for the language model. Finally, the sentiment of an unknown document can be predicted as\vspace{-1mm}
\begin{equation}
    y^i = softmax(W^i f(x) + b^i)\vspace{-1.5mm}
\end{equation}
Where $y^i$ is a predicted sentiment label for a test document, $f$ is a transformer function to encode the input text $x$. We report both the accuracy and weighted f1 scores since the dataset is distributed unevenly across the four labels,

\begin{table}[h!]
\centering
\begin{tabular}{c|ccc} 
     \hline
     PLM & \ac{lstm} & \ac{qlstm} (4q) \\
     \hline
     accuracy & 0.928 & 0.934 \\ 
     weighted f1 & 0.93 & 0.93 \\ 
     \hline
\end{tabular}\vspace{-2mm}
\caption{\label{tab:sa_acc} SA Performance on Multilingual Twitter Dataset}\vspace{-4mm}
\end{table}

Table \ref{tab:sa_acc} summarizes the four-way sentiment classification performance of the local transformer-based classifier initialized with embeddings trained by the classical \ac{lstm} and the quantum \ac{lstm}, while keeping all other hyperparameters the same. The embedding trained by \ac{qlstm} achieves slightly higher accuracy than the classical \ac{lstm}, and has the same weighted f1 score. Our results confirm that the decentralized \ac{qlm} training and the portable information transmission do not sacrifice performance. With the random circuits in its \ac{vqc} gates, the quantum model has higher expressibility - a circuit’s ability to generate states in the Hilbert space \cite{sim2019expressibility}. High expressibility allows the model to better search the solution space given the training data compared to the classical model, despite having the same number of parameters. The high expressibility of the quantum model makes it a ``better learner," contributing to the slightly higher accuracy.

\vspace{-1mm}
\subsection{Ablation Studies}\vspace{-1mm}
\subsubsection{Effect of Number of Qubits}\vspace{-1mm}
A single qubit can be described by a two-dimensional Hilbert space $\mathcal{H}$. Then, a system of $n$ qubits is the tensor product of $n$ such Hilbert spaces\vspace{-3mm}
\begin{equation}
    \underbrace{\mathcal{H} \otimes \mathcal{H} \otimes \cdots \otimes \mathcal{H}}_n\vspace{-4mm}.
\end{equation}\\
We explore the effect of having more qubits in the \ac{vqc} gates to examine the embeddings trained on the higher dimensional Hilbert space. Both the 4-Qubit model and 6-Qubit model converge rapidly as shown in Figure \ref{fig:4v6_all}, so we provide a fine-grained comparison of the training loss for each batch in the first epoch in Figure \ref{fig:4v6_1st}. 
The 6-Qubit language model has a much lower perplexity (Table \ref{tab:perp}) and a faster convergence within the first epoch (Figure \ref{fig:4v6_1st}). This indicates that the 6-Qubit model better captures the non-linear relationship in a higher dimensional embedding space to better fit the data. As the number of qubits increases, so does the noise introduced to the system. However, the smooth loss curve demonstrates model stability as we extend the model to more qubits, indicating the model's robustness against quantum noise.

\vspace{-2mm}
\subsubsection{Effect of Training Data Size}\vspace{-2mm}
\begin{table}[h!]
\centering
\vspace{-2mm}
\begin{tabular}{c|ccc} 
     \hline
     Dataset & Multilingual Twitter & Code-Mixing\\
     Data size & 69491 & 15000 \\ 
     Vocab size & 17000 & 5000 \\ 
     \hline
     perplexity & 1153.672 & 368.031 \\ 
     \hline
\end{tabular}\vspace{-2mm}
\caption{\label{tab:datasize} Model Perplexity for Different Data Sizes}\vspace{-3mm}
\end{table}

We train the \ac{qlstm} on a smaller corpus to confirm that the high perplexity on both the classical and the quantum model is due to the small model size trained on a large dataset. 
The small dataset has a much lower model perplexity (Table \ref{tab:datasize}) and a lower training loss (Figure \ref{fig:datasize}). Due to computation constraints imposed by the \ac{NISQ} machines, our model capacity is limited, so the results indicate parameter saturation but also justify the large perplexity. 
Despite the difference in the corpus size, there is no obvious difference in the convergence speed of the two models. Furthermore, although the code-mixing corpus is more linguistically complex and thus harder to predict, the linguistic complexity still does not overpower the effect of dataset size on the model loss and perplexity. 
Therefore, this experiment demonstrates model stability and justifies the high perplexity due to limited model size.

\vspace{-3mm}
\section{Conclusions}\vspace{-3mm}
This work provides a solution to integrate secure and fast quantum language modeling and flexible classical downstream applications. 
We proposed a decentralized training framework for \ac{qlm}s that provides both parameter protection and ad hoc portability. 
We show that the embedding extraction is sufficient for downstream tasks, which enables the ad hoc transfer of the model from the quantum server to the classical machine.
Our \ac{qlm} achieves competitive if not better results compared to its classical counterpart on multilingual Twitter \ac{sa} tasks. We also studied the effect of the number of qubits in the \ac{vqc} and the training corpus size to demonstrate model stability. 

Our work provides a promising direction and a theoretical foundation for future NLP research that have privacy protection demands. Some of the future work include extending this decentralized quantum training procedure into a wider range of language model architectures. As the required quantum computing hardware becomes available to train large-scale \ac{qlm}s, our approach can be used to easily transmit the quantum information to downstream classical tasks. The ad hoc portability method can also be further studied to enable downstream fine-tuning and model compression.

\clearpage

\begin{footnotesize}
\bibliography{reference}
\end{footnotesize}

\end{document}